# Automated Real-World Sustainability Data Generation from Images of Buildings

Peter J Bentley
*Department of Computer Science*
UCL
London, UK
p.bentley@cs.ucl.ac.uk

Soo Ling Lim
*Department of Computer Science*
UCL
London, UK
s.lim@cs.ucl.ac.uk

Rajat Mathur
*TheSqua.re*
London, UK
rajat@thesqua.re

Sid Narang
*TheSqua.re*
London, UK
sid.narang@thesqua.re

*Abstract*—When data on building features is unavailable, the task of determining how to improve that building in terms of carbon emissions becomes infeasible. We show that from only a set of images, a Large Language Model with appropriate prompt engineering and domain knowledge can successfully estimate a range of building features relevant for sustainability calculations. We compare our novel image-to-data method with a ground truth comprising real building data for 47 apartments and achieve accuracy better than a human performing the same task. We also demonstrate that the method can generate tailored recommendations to the owner on how best to improve their properties and discuss methods to scale the approach.

## I. Introduction

Sustainability is an important consideration in the construction of new buildings today. For example, modern apartment design will often ensure efficiency of insulation, the latest high-efficiency double or triple glazing and community heating schemes where all heating is provided from one efficient centralised boiler, possibly even via solar heating or heat pumps.

But our world is filled with countless existing buildings that may not have been built so recently. To meet global emissions targets and regulations these buildings must be retrofitted with appropriate upgrades and made more sustainable. The first step towards this goal is to identify which are in need of updating and how best to retrofit. For residential dwellings in many countries this is a challenge because of lack of data. Expert surveyors cannot assess every property in the world, and residents or landlords may not have the knowledge to understand which elements of their homes need renewing.

We address this challenge in this work by introducing a new image-to-data approach for sustainability data generation from photographs of buildings. We show for the first time that the latest generative artificial intelligence models can be used in combination with domain-specific knowledge to derive data that is sufficient to enable the calculation of features such as carbon footprint and tailored recommendations to owners on how to improve the efficiency of their properties.

The rest of the paper is organised as follows: the next section provides a background of related work in this area. Section III describes our approach. Section IV provides our validation through empirical testing. We conclude in Section V.

## II. Background

### A. Sustainability and Property Data

With the right data it is possible to understand and improve how consumers make more sustainable choices [1]. But property data is difficult to obtain. In the UK, Energy Performance Certificates provide useful data on the construction and efficiency of properties. But EPCs are frequently not available for all properties, their data (especially ratings) [2] may be inconsistent over time, and EPCs are not a global solution – only the UK and a few EU countries produce such certification and make it available [3]. Some studies examine the profitability of energy efficiency retrofitting work. For example, [4] proposes a rental housing framework (RentalCal) to calculate profitability of energy efficiency retrofitting work in Europe. Sustainable profits resulting from retrofits is calculated using energy-demand modeling combined with retrofit option rankings and life-cycle cost analysis [5]. While many focus on $CO_2$ to provide a metric of sustainability, even this is called into question by some researchers, who claim that the idea was first proposed in a 'successful, deceptive' marketing campaign from an oil company [6] and more holistic notions such as 'carbon shadow' [7] may provide a more nuanced method of understanding environmental impact. In this work we focus on identification of measurable property features that are known to be correlated to efficiency: e.g., glazing, lighting, heating and building age.

### B. Large Language Models (LLM)

With data so sparse, one option used by some is to use AI (Large Language Models) or agent based learning to generate training data for models [8] [9]. LLMs have been used widely for architectural applications, e.g. even in the generation of new building designs [10]. In our previous work, we also developed a data-driven address-specific metric called EcoGrade, which integrates government approved datasets and uses interpolation where data is sparse [3]. But interpolation is only possible when there is sufficient similar data. Where there is simply no data at all, a different approach is required. While LLMs are now beginning to be used for tasks such as chart understanding by converting data-filled visuals into textual summaries [11] there is little work using LLMs to perform image-to-data conversion in this area: starting with images of buildings and generating usable sustainability data.

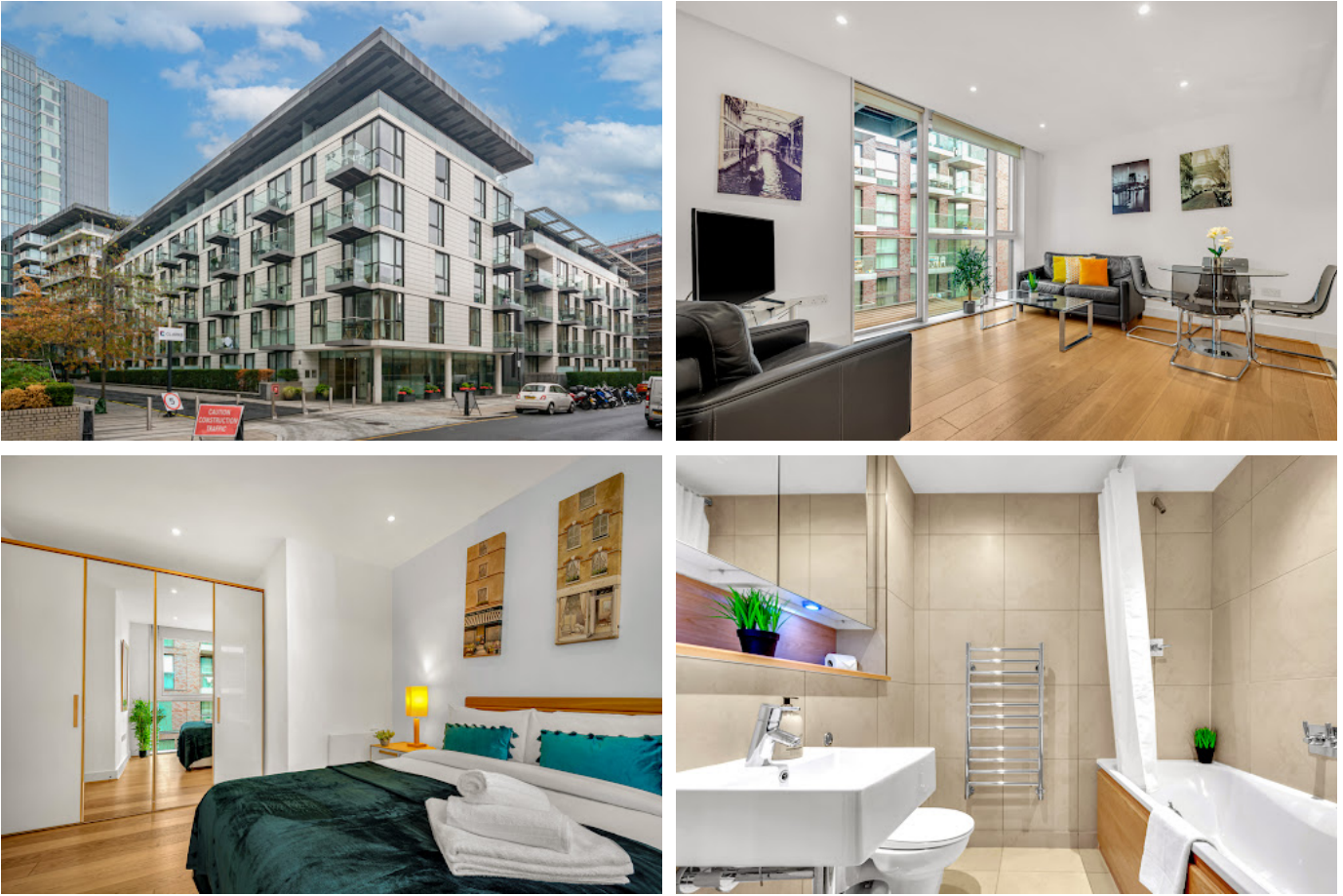

Fig. 1 Example images from the real-world data set. Top left shows an external view of a modern apartment block. Top right shows wide-angle living room illustrating windows with view of neighbourhood, and lighting. Bottom left shows wide-angle of bedroom, illustrating heating source (electric panel next to light), lighting and further reflected views of window and neighbourhood. Bottom right shows bathroom with electric towel radiator.

## III. METHOD

This work is performed in collaboration with a global accommodation marketplace TheSqua.re. With their support we are able to prepare a ground truth dataset of 46 apartments each with confirmed data. Our first step was the preparation of this unique dataset. We then describe the sustainable data-generation algorithm which combines domain knowledge with prompt engineering of GPT-4o to analyse images correctly and derive the correct values according to this ground truth.

### A. TheSquare

TheSqua.re is a digital corporate housing and luxury alternate accommodation marketplace with 200,000 furnished apartments managed by 2000 plus operators in 600 cities globally. They provide a large choice of alternate accommodations in major cities, using proprietary technology. In 2020, they launched MySqua.re, a private label brand that delivers a portfolio of homes in London offering city centre locations, currently live in more than ten neighbourhoods in London such as Fitzrovia, Mayfair, Kensington, Canary Wharf and City of London, operating more than 100 apartments. TheSqua.re Group (includes TheSqua.re and MySqua.re) aims to enable its customers (both building suppliers and renters) to transition to more sustainable practices.

TheSqua.re has provided the data for the MySqua.re apartments, which includes Energy Performance Certificates that detail the type of windows, main heating system, and other useful information. They also provide the listing photographs used on their website: typically a set of images (from 10 to 60 per property) that show the interior of the rooms and sometimes exterior views of the building, e.g. see Fig. 1.

TABLE I. INITIAL REAL-WORLD DATA PROVIDED BY INDUSTRY PARTNER (47 UNIQUE PROPERTIES)

| Address | EPC URL | Building age | images |
|---|---|---|---|
| apartment number, building name, street address, postcode | https://find-energy-certificate.service.gov.uk/energy-certificate/xxx-xxx-xxx-xxx-xxx | low-high | image1URL; image2URL; image3URL; … image*n*URL |

TABLE II. GROUND TRUTH DATASET

| ID | [A] | [B] | [C] | [D] | [c1] | [c2] | [c3] | [c4] | [c5] | [c6] | [c7] |
|---|---|---|---|---|---|---|---|---|---|---|---|
| Feature | Building images | Heating images | Window images | Lighting images | Building age | Building type | Main Heating source | Window type | Lighting | Energy Kwh/m2 | EPC rating |
| 0 | 0;2;3;5 | 1;5;8 | 0;2;3;4;5 | 1;2;5 | 2014 | >5 units | Community scheme with underfloor heating | High efficiency double glazed | Low energy in 75% | 57 | B |
| … | … | … | … | … | … | … | … | … | … | … | … |
| 46 | 0;1;12 | 3;5;7 | 7;9 | 0;3;15 | Before 1900 | 2-4 units | Gas boiler water radiators | Single glazed | Low energy in 100% | 275 | D |

## B. Ground Truth Data Preparation

While the data provided by our collaborator is extensive, it required careful processing to be useful for our purposes. Table 1 shows the data provided. Table II shows the features we extracted for our ground truth. An approximate building age was provided for around a third of the apartments; we investigated each building using their addresses and discovered precise building construction dates for all by finding planning permission documents on government websites, developer's brochures, or using legacy sales listings of many years ago that were described as "new build". The same data sources enabled us to determine the building type, for which we use the EIA (US Energy Information Adminiistration) categorization: (1) Single-family detached; (2) Single-family attached; (3) Apartments in buildings with 2–4 units; (4) Apartments in buildings with 5 or more units; (5) Mobile home. Window type, main heating source, lighting, energy and EPC rating were all obtained from the EPC associated with the apartment. In several cases however the apartments had no EPC, in which case these data were obtained from legacy sales documents and plans found online. In the few cases where no precise data was available, the values were estimated based on similar properties built nearby at the same time.

For each category of data we select 1 to 5 images that may be indicative of the correct value, e.g. for building age we select an image of the building exterior where available and images showing views from windows; for window type we select images of windows; for heating we select wide-angle images of interiors that should include radiators if they exist; for lighting we select wide-angle images of interiors that should include lights, see Fig. 1 for example images. While this image categorization and selection was performed manually here, the addition of another stage using GPT-4o (or another simpler image-to-text model) to perform this process automatically would be completely feasible in future work.

## C. Efficiency Data Generation

The task we need to perform is for every building ID, to convert the selected images (columns [A] to [D] in Table II) into the efficiency data (columns [c1] to [c7] of Table II). Our ground truth enables us to assess the accuracy of the approach; in normal usage there will be no ground truth and only the photographs will be available.

While today's large language models are remarkable, they do not have the specific domain knowledge necessary to perform our desired function without support. Prompt GPT-4o to simply find desired values for our photos and it becomes confused, frequently making "intelligent guesses" based on its general knowledge while missing the subtle clues within the photos. Air vents are confused with lights; building features are interpreted incorrectly. Thus to make best use of this generative model, we need to give it sufficient knowledge within the prompts in order to guide its focus within the images. We also need to take certain elements of reasoning and decision-making away from the AI as they confused its attention, instead asking it to focus more on whether it could identify specific elements within images, and not ask it to interpret the meaning of those elements – which it lacked the knowledge to perform correctly. This makes use of the strongest capabilities of the model (image to text conversion) and keeps the algorithm logic consistent and reliable. Finally, it was necessary to break the task into a series of separate requests to the LLM, each with its own prompt and images, and in some prompts it was necessary to ask for a short explanation before giving its answer. This simple "chain of thought" prompting (give initial question, enable it to break down the problem into smaller steps, and then provide the answer) helps the LLM to identify and incorporate multiple factors that might contribute to the final answer. Our preliminary experiments showed this was necessary to stop the LLM from "jumping to conclusions" – giving a quick answer which might ignore relevant factors.

The algorithm is illustrated in Fig. 2 and Tables III and IV. The overall task is split into five steps, each using a different prompt and image set: (1) building age; (2) building type; (3)

heating type and energy; (4) window type; (5) lighting efficiency. The third step to identify heating is the most complex. We require specific features to be identified and then a set of domain knowledge rules to be followed to draw conclusions about the heating source. We found the most reliable method (and cost effective in terms of reducing prompt length) is to ask the LLM to perform the identification, returning a list of heating methods and then use ordinary code to apply the knowledge base rules (Table III) which also use results from the previous two steps to make the final determination of heating type and source energy.

### D. Energy Consumption Data Generation and Automatic Recommendations

While the focus of this work is on the automatic estimation of energy efficiency of apartments, we also attempt the automated estimation of energy consumption in kwh/m$^2$ (from which $CO_2$ can be calculated when heating energy source is known [3]). To achieve this we provide the LLM with all results from prompts P1 to P5 plus the results from the application of our domain knowledge rule base, together with Building Images [A]. We use the prompt P6 as shown in Table IV. While it is implausible to achieve perfect results, even a rough estimation may be better than nothing. Finally we investigate whether the LLM can generate recommendations on how best to upgrade the apartments, providing the using the same input data, see prompt P7, Table IV.

## IV. EXPERIMENTS AND RESULTS

The methods described above were implemented in Python using Jupyter notebook, "model": "gpt-4o" with default settings. We perform three experiments: (1) we run the efficiency data generation method and (2) we run the energy consumption data generation method as described above. We compare results with ground truth values and also compare results with those from a human volunteer with more than 10 years experience in the UK property market (shown identical images as given to the AI). Finally, (3) we explore whether the LLM can provide useful recommendations to the owner of the apartment when given the text resulting from P1 to P5, using prompt P7.

Table V provides the results. Building age was on average 4.47 years incorrect by the AI (for ages spanning more than 100 years). Building type predicted by the AI was correct in every example but one. Heating type and energy source were correctly predicted in 76.6% and 74.4% instances respectively. Window types were estimated perfectly for 53% of the apartments plus approximately correct (double glazing instead of high efficiency double glazing or vice versa) for another 42.6% making 95.6% in total. Lighting efficiency (0 to 100% low efficiency) achieved an RMSE of 27.8%. Average difference of energy consumption (in a range from zero to 450+ kwh/m$^2$) was on average wrong by 42.66 kwh/m$^2$. Although not perfect, the AI did better than our experienced human volunteer in every category except lighting. Finally, recommendations were successfully generated (Table IV bottom right) that provided sensible advice.

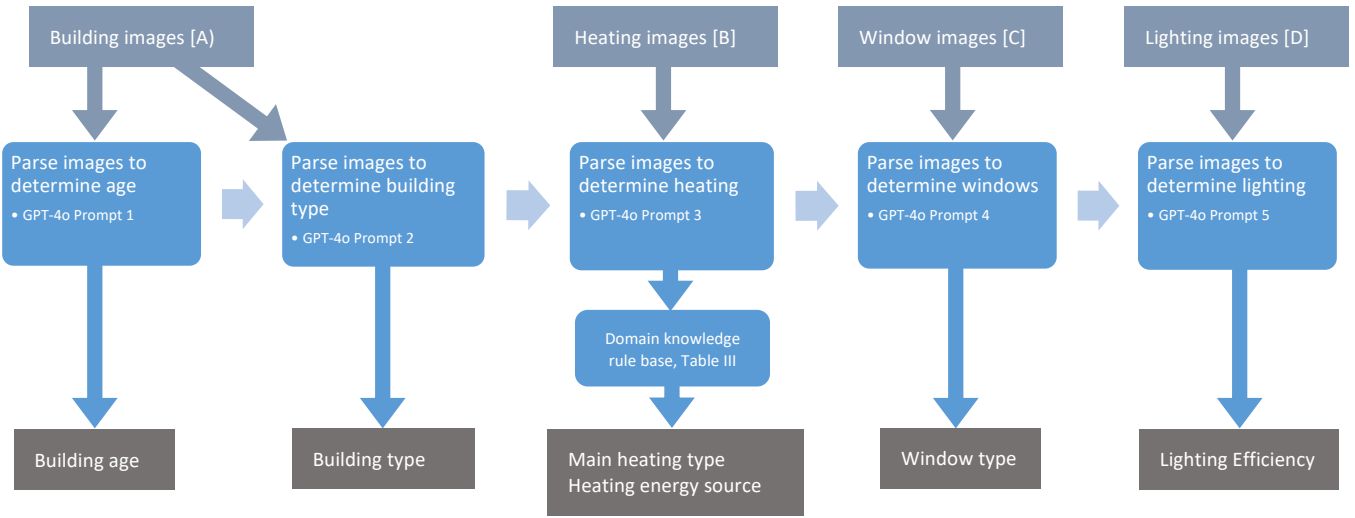

Fig. 2 Efficiency data generation from building images flow diagram.

TABLE III. DOMAIN KNOWLEDGE RULE BASE USED TO CONVERT GPT-4O OBSERVATIONS INTO LIKELY MAIN HEATING TYPE AND ENERGY SOURCE.

| Main Heating type | Heating Energy Source |
|---|---|
| if radiators == 'Y' and water_filled == 'Y':  main_heating = 'water rads' <br> elseif panel == 'Y':  main_heating = 'electric panels' <br> elseif storage == 'Y':  main_heating = 'electric storage' <br> if air_vent == 'N' and radiators == 'N' and panel == 'N' and storage == 'N':  main_heating = 'underfloor' <br> if air_vent == 'Y' and radiators == 'N' and panel == 'N' and storage == 'N':  main_heating = 'warm air' | If Building age >= 1970 and Building Type is > 5 units and panel == 'N' and storage == 'N': energy_source = 'community' <br> elseif water_filled == 'Y': energy_source = 'gas' <br> elseif heating_type == 'underfloor' or panel == 'Y' or storage == 'Y': energy_source = 'electric' |

TABLE IV. PROMPTS USED TO INSTRUCT GPT-4O TO PERFORM THE DESIRED FUNCTION WITH EXAMPLE OUTPUT FOR ONE APARTMENT.

| No. | Prompt text | Example GPT-4o output for one apartment |
|---|---|---|
| P1 | The images below belong to the same apartment. The building is located in UK. What is the age of this apartment?<br><br>If the image is an internal shot, look carefully at the architectural features or characteristics of the room, such as the ceiling height, whether or not the ceiling is coved or has ceiling roses, the floor material, the type of windows and doors. If there is a window in the image, you may base your judgement on the view out the window. If the image is an external shot, look at the building itself. In the UK brick may be used during any period, but older buildings are unlikely to be towers or use large amounts of glass and steel.<br><br>First provide a short explanation, then select one of these options: (1) before 1900, (2) 1900-1930, (3) 1930-1950, (4) 1950-1970, (5) 1970-1990, (6) 1990-2020, (7) 2020-now. | Based on the architectural features and design of the apartment shown in the images, it is evident that this building has very modern characteristics. The large windows, open plan layouts, and contemporary furnishings are indicative of recent trends in architecture. The windows are large and allow for significant natural light, which is common in more recent constructions. Additionally, the finishes and materials, such as the flooring and lighting, align with modern design aesthetics.<br><br>Given these observations, the building most likely belongs to the most recent planning and construction trends. Therefore, the age of this apartment is best classified as: 2020-now. |
| P2 | The images below belong to the same apartment. The building is located in UK. What is the housing unit type of this apartment?<br><br>If the image is an internal shot, get clues from the characteristics of the room, the type of windows and doors. If there is a window in the image, you may base your judgement on the view out the window. If the image is an external shot, look at the building itself.<br><br>First provide a short explanation, then select one of these options: (1) Single-family detached (2) Single-family attached (3) Apartments in buildings with 2–4 units (4) Apartments in buildings with 5 or more units (5) Mobile home | Based on the provided images, the style of windows, views of surrounding buildings and cityscape indicate that this apartment is likely located in a multi-story building within an urban environment. The characteristics and layout of the rooms also suggest a modern apartment setting in a densely populated area.<br><br>Thus, the housing unit type of this apartment is most consistent with: (4) Apartments in buildings with 5 or more units |
| P3 | The images below belong to the same apartment. The building is located in UK. Tell me what heating systems you can find in the images.<br><br>Can you see anything in the ceiling resembling a long air vent? Can you see radiators? If so, are they water filled? Water filled radiators will have pipes or thermostatic valves. Can you see electric panel heaters? Electric radiators are often flat panels and may have electric cables to a wall outlet. Can you see electric storage heaters? Storage heaters are often flat with vents and may have electric cables to a wall outlet.<br><br>Your answer should look like this: {{ "Air vent": "Y/N", "Radiators": "Y/N", "Water filled": "Y/N", "Electric panel heaters": "Y/N", "Electric storage heaters": "Y/N" }} | Row Number: 0<br>Raw: { "Air vent": "Y", "Radiators": "N", "Water filled": "N", "Electric panel heaters": "N", "Electric storage heaters": "N" } |
| P4 | The images below belong to the same apartment. The building is located in UK. What type of windows does this apartment have?<br><br>Single glazed windows are likely to be in the oldest properties and may by characterized by distinctive metal handles used to open them. The efficiency of double glazed windows will be better for more modern buildings.<br><br>First provide a short explanation, then select one of these options: (1) single glazed (2) double glazed (3) high efficiency double or triple glazed | The windows in the apartment shown in the images appear to have a modern design with thick framing. The construction and style suggest they are more energy-efficient than older single glazed windows, which usually feature metallic handles and simpler designs.<br><br>(3) high efficiency double or triple glazed |
| P5 | The images below belong to the same apartment. The building is located in UK. What type of lighting does this apartment have?<br><br>The more recently the apartment has been decorated, the more energy efficient the lighting is likely to be. Lighting efficiency may be slightly reduced if you see older-looking free-standing lamps, table lamps, wall lighting or pendant lighting.<br><br>First provide a short explanation, then select one of these options: (1) no low energy lighting (2) low energy in 20% (3) low energy in 40% (4) low energy in 60% (5) low energy in 80% (6) low energy in 100% | The images show a mix of floor and table lamps as well as ceiling recessed lights. Recessed lights typically use LED which is energy-efficient, while the actual bulbs in the other lamps are not visible. Based on the visible energy-efficient lighting fixtures (like the recessed ceiling lights) and the assumption that this modern-looking apartment likely uses some energy-efficient bulbs in other lamps as well, it is reasonable to conclude there is a significant portion of low-energy lighting. Therefore, the apartment likely has: low energy in 80% |
| P6 | The images below belong to the same apartment. The building is located in UK. Estimate the energy consumption in kwh per metre squared for the following apartment.<br><br>The following details describe the apartment shown in the images. [P1 to P5 output]<br><br>First provide a short explanation, then give an estimate of the kwh. A new and highly efficient apartment might have a kwh/m2 value as low as 35 or better. An older, inefficient apartment might have a kwh/m2 value as high as 450 or worse. | Considering these high-efficiency characteristics, the energy consumption is expected to be quite low. For a new and highly efficient apartment, the energy consumption per square meter can be as low as 35 kWh/m² per year. Therefore, the energy consumption for this apartment is estimated to be around 35 kWh/m² to 50 kWh/m². |
| P7 | The images below belong to the same apartment. The building is located in UK.<br><br>The following details describe the apartment shown in the images. [P1 to P5 output]<br><br>Based on the images and the details, generate a short recommendation to the owner on how to improve the property's energy efficiency. | Based on our observations, your beautifully renovated Victorian-era apartment already has several modern features in place. However, there are further enhancements you can implement to maximize its energy efficiency. Consider the following recommendations:<br><br>1. **Window Upgrades:** - **Improve Glazing:** While your current double-glazed windows are more efficient than single glazing, upgrading to triple glazing can further enhance thermal insulation and reduce heat loss […cut for brevity…] |

TABLE V. RESULTS FROM EXPERIMENTS

| | Age Av error (years) | Building Type (% correct) | Heating type (% correct) | Energy source (% correct) | Window type perfect+approx (%) | Lighting Eff RMSE (% low energy) | Energy consump Av diff (kwh/m²) |
|---|---|---|---|---|---|---|---|
| AI | 4.47 | 98 | 76.6 | 74.4 | p53 a42.6 | 27.82 | 42.66 |
| Human | 5.87 | 91.5 | 72.34 | 38.3 | p61.7 a31.9 | 34.18 | 60.39 |

## V. DISCUSSION

The results have demonstrated that it is possible to generate apartment efficiency and energy consumption estimates with a surprising level of accuracy from images alone. These values can be used to provide sustainability scores to people, and could help provide recommendations on how to improve properties as shown. However it could be argued that another approach might be possible. In EPC reports, the efficiency of apartments is measured in an overall rating from A to G (Table II [c7]). Perhaps the LLM could estimate this rating from the textual output of prompts P1 to P5. Or perhaps the LLM could directly estimate the rating from images alone. Two new prompts were created (text too long to include here) to test this idea, which (a) instructed the LLM to estimate the EPC rating based on P1 to P5 text; and (b) provided Building Images [A] and asked for the EPC rating. Results were revealing: (a) using textual input alone achieved scores with RMSE of 1.088 – on average slightly more than one letter wrong (e.g. A for a B; E for a C). The cost for calling the API was also greater. Results for (b) using Building Images [A] achieved an RMSE of 0.858 – on average less than one letter wrong. It is therefore clear that images paint a thousand words (or at least more words than the output for P1 to P5). Despite this finding we do not recommend the use of AI to generate EPC ratings: these scores are inconsistent in real EPC reports and vary by country and by year as governments change policies, which then alter the calculation method used. We argue our approach is better: identify clear features within dwellings and use the physical evidence to help provide advice to owners.

The creation of this method required four stages of human input, but many of these could be automated in future: First the selection of appropriate images per feature could be easily automated. Second, the writing of prompts containing clues on which elements are important (e.g. "high ceilings, radiators with water pipes") could in theory be derived by an AI that iteratively fine-tunes prompts by using the results of machine learning to correlate textual features in ground truth data with mistaken features in AI output. Third, the creation of the knowledge base (Table III) could be replaced by fine-tuning of the AI with domain-specific data such that it learns how best to interpret the features it discovers. Fourth, while prompt engineering will probably always be required to ensure LLMs perform appropriately, future models may be better able to perform complex tasks such as feature identification and complex domain-specific reasoning about those features, without the need for manually breaking the task into several separate prompts and image sets as was done here. Finally, while the recommendations provided for owners were sensible and correct, the use of fine-tuning with knowledge of local options would make recommendations even more pertinent.

## VI. CONCLUSIONS

There will never be accurate data that describes the design and construction of the majority of buildings that currently exist in the world today. These buildings may be recently built or may have been standing for centuries. Yet our need to improve efficiency of all buildings demands such data. In this work we demonstrate how the latest image to text LLMs when combined with domain knowledge and when appropriate prompt engineering is employed can generate useful and relatively accurate estimations of building features and even energy consumption compared to ground truth of real-world data. Our image-to-data approach performs better than an experienced human performing the same task. It also is capable of providing recommendations for owners on how best to improve their properties. While this work is a clear proof of concept we anticipate that further improvements would enable the method to be scaled up to properties worldwide.